\documentclass[10pt,twocolumn,letterpaper]{article}

\usepackage{cvpr}              

\usepackage{graphicx}
\usepackage{amsmath}
\usepackage{amssymb}
\usepackage{booktabs}
\usepackage{color}

\usepackage[pagebackref,breaklinks,colorlinks]{hyperref}

\usepackage[capitalize]{cleveref}
\crefname{section}{Sec.}{Secs.}
\Crefname{section}{Section}{Sections}
\Crefname{table}{Table}{Tables}
\crefname{table}{Tab.}{Tabs.}

\def\cvprPaperID{5134} 
\def\confName{CVPR}
\def\confYear{2022}

\usepackage{setspace}
\usepackage{bm}
\usepackage{multirow}
\usepackage{pifont}
\usepackage{balance}
\usepackage{makecell}
\usepackage{colortbl} 
\usepackage{cuted} 
\definecolor{mygray}{gray}{.9}
\usepackage{hyperref}

\begin{document}

\title{Learning What Not to Segment: A New Perspective on Few-Shot Segmentation}

\author{Chunbo Lang
\hspace{0.6cm}
Gong Cheng\thanks{Gong Cheng is the corresponding author.}
\hspace{0.6cm}
Binfei Tu
\hspace{0.6cm}
Junwei Han
\vspace{0.0cm}\\
{\large School of Automation, Northwestern Polytechnical University, Xi'an, China}
\vspace{0.00cm}\\
{\tt\small \{langchunbo,\,binfeitu\}@mail.nwpu.edu.cn,}\,
{\tt\small \{gcheng,\,jhan\}@nwpu.edu.cn}

}
\maketitle

\begin{abstract}
   Recently few-shot segmentation (FSS) has been extensively developed. Most previous works strive to achieve generalization through the meta-learning framework derived from classification tasks; however, the trained models are biased towards the seen classes instead of being ideally class-agnostic, thus hindering the recognition of new concepts. This paper proposes a fresh and straightforward insight to alleviate the problem. Specifically, we apply an additional branch (base learner) to the conventional FSS model (meta learner) to explicitly identify the targets of base classes, \textup{i.e.}, the regions that do not need to be segmented. Then, the coarse results output by these two learners in parallel are adaptively integrated to yield precise segmentation prediction. Considering the sensitivity of meta learner, we further introduce an adjustment factor to estimate the scene differences between the input image pairs for facilitating the model ensemble forecasting. The substantial performance gains on PASCAL-5$^i$ and COCO-20$^i$ verify the effectiveness, and surprisingly, our versatile scheme sets a new state-of-the-art even with two plain learners. Moreover, in light of the unique nature of the proposed approach, we also extend it to a more realistic but challenging setting, \textup{i.e.}, generalized FSS, where the pixels of both base and novel classes are required to be determined. The source code is available at \href{https://github.com/chunbolang/BAM}{github.com/chunbolang/BAM}. 
\end{abstract}

\section{Introduction}
\label{sec:1}
Benefiting from the well-established large-scale datasets \cite{deng2009imagenet,everingham2010pascal,lin2014microsoft}, a wealth of convolutional neural network (CNN) based computer vision techniques have undergone rapid development over the past few years \cite{ren2015faster,redmon2016you,lin2017feature,lin2017focal,long2015fully,ronneberger2015u,simonyan2014very,he2016deep,huang2017densely,he2017mask}. However, collecting sufficient labeled data is notoriously time-consuming and labor-intensive, especially for dense prediction tasks, such as instance segmentation \cite{he2017mask,bolya2019yolact,bolya2019yolact++,xie2020polarmask,lee2020centermask} and semantic segmentation \cite{long2015fully,ronneberger2015u,lin2016scribblesup,badrinarayanan2017segnet,peng2017large}. In striking contrast with the machine learning paradigms, humans can easily recognize new concepts or patterns from a handful of examples, which greatly stimulates the research interest of the community \cite{vilalta2002perspective,nichol2018first,vanschoren2018meta}. Thus, few-shot learning (FSL) is proposed to address this problem by building a network that can be generalized to unseen domains with scarce annotated samples available \cite{vinyals2016matching,wang2016learning,ravi2016optimization,cheng2021task}.
\begin{figure}[t]
	\centering
	\includegraphics[width=1.0\linewidth]{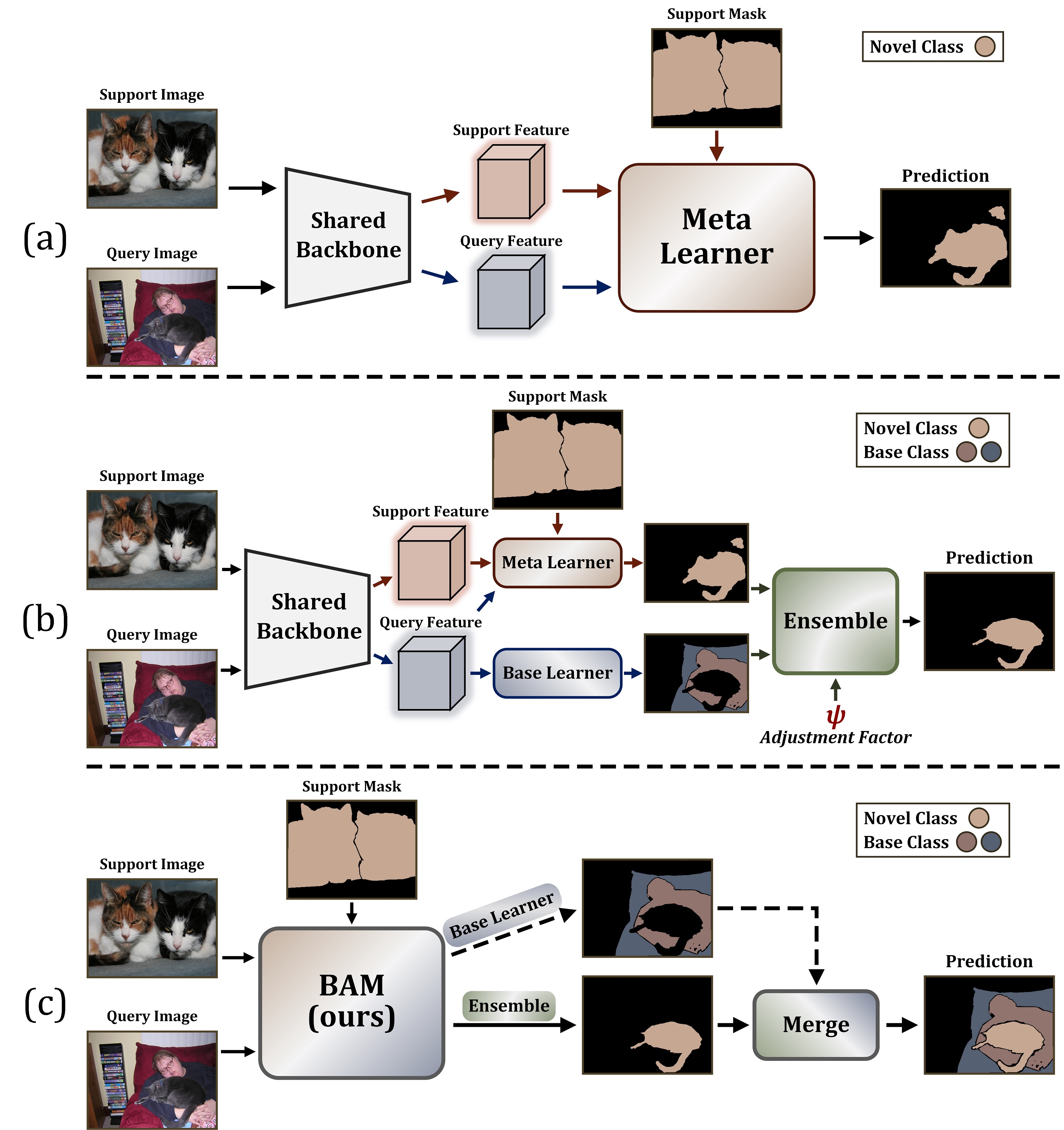}
	\caption{Comparison of our BAM and previous work. (a) Conventional approaches typically employ meta-learning frameworks to train the FSS models, which is inevitably biased towards base classes rather than being ideally class-agnostic, thus hindering the recognition of target objects for novel class (e.g., \textit{cat} ($\vcenter{\hbox{\LARGE$\textcolor[RGB]{198,167,146}{\bullet}$}}$)). (b) Our BAM introduces an additional branch, namely base learner, to explicitly predict the regions of base classes. In this way, the distractor objects (e.g., \textit{person}  ($\vcenter{\hbox{\LARGE$\textcolor[RGB]{145,116,110}{\bullet}$}}$) and \textit{sofa} ($\vcenter{\hbox{\LARGE$\textcolor[RGB]{85,96,114}{\bullet}$}}$)) in the query image can be suppressed significantly after the ensemble module. (c) Extension of our BAM under the generalized FSS settings, where the pixels of both base and novel classes are required to be determined. The refined results are again merged with the output of the base learner to generate comprehensive predictions.}
	\label{fig:1}
\end{figure}\\
\indent In this paper, we undertake the application of FSL in the field of semantic segmentation, termed as few-shot segmentation (FSS), where the model leverages only very \textit{few} labeled training data to segment the targets of a specific semantic category from the raw image \cite{shaban2017one}. Fueled by the success of few-shot classification, most existing FSS approaches strive to achieve generalization through meta-learning frameworks \cite{zhang2018sg,siam2019amp,nguyen2019feature,zhang2019canet,zhang2019pyramid,liu2020crnet,wang2019panet,yang2020prototype,liu2020prototype,liu2020part,wang2020few,li2021adaptive,liu2021anti,zhang2021self,lu2021simpler,min2021hypercorrelation,yang2021mining,wu2021learning,liu2021harmonic}. A series of learning tasks (episodes) are sampled from the base dataset to mimic the few-shot scenarios of novel classes, \textit{i.e.}, match training and testing conditions. However, it is woefully inadequate and underpowered. Meta-training on the base dataset with abundant annotated samples inevitably introduces a bias towards the seen classes rather than being ideally class-agnostic, thus hindering the recognition of new concepts \cite{fan2021generalized}. Notably, when countering hard query samples that share similar categories with base data, the generalization performance might be on the verge of collapse. \\
\indent We argue that aside from designing more powerful feature extraction modules \cite{yang2020prototype,li2021adaptive,xie2021scale}, adjusting the use of base datasets containing sufficient training samples is also an alternative method to alleviate the above-mentioned bias problem, which has been neglected in previous works. To this end, we introduce an additional branch (base learner) to the conventional FSS model (meta learner) to explicitly predict the targets of base classes (see Fig.\,\ref{fig:1}). Then, the coarse results output by these two learners in parallel are adaptively integrated to generate accurate predictions. The central insight behind such an operation is to identify confusable regions in the query image through a high-capacity segmentation model trained within the traditional paradigm, further facilitating the recognition of novel objects. Incidentally, the proposed scheme is named BAM as it consists of two unique learners, \textit{i.e.}, \textbf{b}ase \textbf{a}nd the \textbf{m}eta.\\
\indent Moreover, we notice that meta learners are typically sensitive to the quality of support images, and the large variances between the input image pairs could lead to severe performance degradation. On the contrary, base learners tend to provide highly reliable segmentation results and stable performance due to the single query image as input. Based on this observation, we further propose to leverage the evaluation results of the scene differences between query-support image pairs to adjust the coarse predictions derived from meta learners. Inspired by the style loss that is extensively adopted in the domain of image style transfer \cite{gatys2015neural,gatys2016image,jing2019neural}, we first calculate the difference of the Gram matrices of the two input images and then utilize the Frobenius norm to obtain the overall indicator for guiding the adjustment process. As illustrated in Fig.\,\ref{fig:1}(b), the distractor objects of base classes (e.g., \textit{person} and \textit{sofa}) in the query image are suppressed significantly after the ensemble module, achieving accurate localization of novel objects (e.g., \textit{cat}). Furthermore, in light of the unique character of the proposed approach, we also extend the current task to a more realistic but challenging setting (\textit{i.e.}, generalized FSS), where the pixels of both base and novel classes are required to be determined, as presented in Fig.\,\ref{fig:1}(c). To sum up, our primary contributions can be concluded as follows:

$\bullet\;$ We propose a simple but efficient scheme to address the bias problem by introducing an additional branch to explicitly predict the regions of base classes in the query images, which sheds light on future works. 

$\bullet\;$ We propose to estimate the scene differences between the query-support image pairs through the Gram matrix for mitigating the adverse effects caused by the sensitivity of meta learner.

$\bullet\;$ Our versatile scheme sets new state-of-the-arts on FSS benchmarks across all settings, even with two plain learners.

$\bullet\;$ We extend the proposed approach to a more challenging setting, \textit{i.e.}, generalized FSS, which simultaneously identifies the targets of base and novel classes. 

\section{Related Works}
\label{sec:2}
\textbf{Semantic Segmentation.} Semantic segmentation is a fundamental computer vision task that aims to recognize each pixel of the given images according to a set of predefined semantic categories \cite{ronneberger2015u}. Recently, tremendous progress has been made in this field benefited from the advantages of fully convolutional networks (FCNs) \cite{long2015fully}. Various robust network designs have been proposed successively, also bringing with them some fundamental techniques, such as dilated convolution \cite{yu2015multi}, encoder-decoder structure \cite{ronneberger2015u}, multi-level feature aggregation \cite{lin2017refinenet}, attention mechanism \cite{huang2019ccnet}, etc. However, conventional segmentation models require sufficient annotated samples to produce satisfactory results and hardly generalize to unseen categories without fine-tuning, thereby hindering to some extent their practical applications. In this work, atrous spatial pyramid pooling (ASPP) module \cite{chen2017deeplab} based on dilated convolution is introduced into the meta learner to enlarge the receptive filed, and PSPNet \cite{zhao2017pyramid} is served as the base learner to predict the distractor objects of base categories. 

\textbf{Few-Shot Learning.} The computer vision community has made ongoing efforts over the years to render a network with the ability to generalize to novel categories. Most current methods in the few-shot learning (FSL) domain follow the meta-learning framework proposed in \cite{vinyals2016matching}, where a set of learning tasks (episodes) are sampled from the base dataset to mimic the few-shot scenarios. On this basis, FSL approaches can be further subdivided into three branches: (i) metric-based \cite{snell2017prototypical,sung2018learning,li2019finding}, (ii) optimization-based \cite{ravi2016optimization,finn2017model,jamal2019task}, and (iii) augmentation-based \cite{chen2019imaged,chen2019imageb}. Our work is closely related to the metric-based approach that determines the affinity between the support prototypes \cite{snell2017prototypical} and query features with a specific distance measure, such as Euclidean distance and cosine distance. Inspired by the generalized setting in FSL \cite{li2019few}, we attempt to help recognize new targets by predicting the regions of base classes in the query images, and the segmentation task in a low-data regime is also extended to this setting.

\textbf{Few-Shot Segmentation.} Few-shot segmentation (FSS) is a natural application of the FSL technique for dense prediction tasks, and it has received increased attention in recent years. Previous approaches typically employ a two-branch structure, \textit{i.e.}, support branch and query branch, to transfer annotation information and interact between extracted features. Shaban \textit{et al}. \cite{shaban2017one} proposed the pioneering work in this field, termed OSLSM, in which the support (conditional) branch is utilized to generate the classifier weights for query branch prediction. Later on, Zhang \textit{et al}. \cite{zhang2018sg} exploited the masked average pooling operation to obtain representative support features, which also served as the fundamental technology for subsequent works. More recently, some relevant research abandoned the training process of heavy backbone networks in favor of building powerful blocks on the fixed ones to boost performance, such as CANet \cite{zhang2019canet}, PFENet \cite{tian2020prior}, ASGNet \cite{li2021adaptive}, SAGNN \cite{xie2021scale}, and MM-Net \cite{wu2021learning}. 

However, the generalization performance of these methods heavily depends on the meta-learning framework, which could be fragile even with the fine-tuning process. More specifically, the trained FSS models are biased towards base classes due to the unbalanced data distribution and large domain shift. We observe that very few works in this field explicitly study the generalization degradation problem but focus on designing high-capacity interaction modules between the two branches. Tian \textit{et al}. \cite{tian2020prior} leverages the high-level features extracted from the fixed backbone network to evaluate the similarity, providing important segmentation cues for the query images. Such a parameter-free approach could help the network learn to capture more generic patterns, thereby improving generalization. Instead, this paper concentrates on a more fundamental perspective to address the bias problem by explicitly identifying the confusable regions of base classes. 

\section{Problem Definition}
Few-shot segmentation aims at performing segmentation with only a few labeled data. Current approaches typically train models within the meta-learning paradigm, also known as episodic training. Specifically, given two image sets $\mathcal{D}_{{\rm{train}}}$ and $\mathcal{D}_{{\rm{test}}}$ that are disjoint in terms of object categories, the models are expected to learn transferable knowledge on $\mathcal{D}_{{\rm{train}}}$ with sufficient annotated samples and thus exhibit good generalization on $\mathcal{D}_{{\rm{test}}}$ with scarce annotated examples. In particular, both sets are composed of numerous episodes, each of which contains a small support set $\mathcal{S} = \left\{ {\left( {{\bf{x}}_i^{\rm{s}},{\bf{m}}_i^{\rm{s}}} \right)} \right\}_{i = 1}^K$ and a query set $\mathcal{Q} = \left\{ {\left( {{{\bf{x}}^{\rm{q}}},{{\bf{m}}^{\rm{q}}}} \right)} \right\}$, where ${{\bf{x}}^ * }$ and ${{\bf{m}}^ * }$  represent a raw image and its corresponding binary mask for a specific category $c$, respectively. The models are optimized during each training episode to make predictions for the query image ${{\bf{x}}^{\rm{q}}}$ under the condition of the support set $\mathcal{S}$. Once the training is complete, we will evaluate their few-shot segmentation performance on ${\mathcal{D}_{{\rm{test}}}}$ across all the test episodes, without further optimization.

\begin{figure*}[t]
	\centering
	\includegraphics[width=0.95\linewidth]{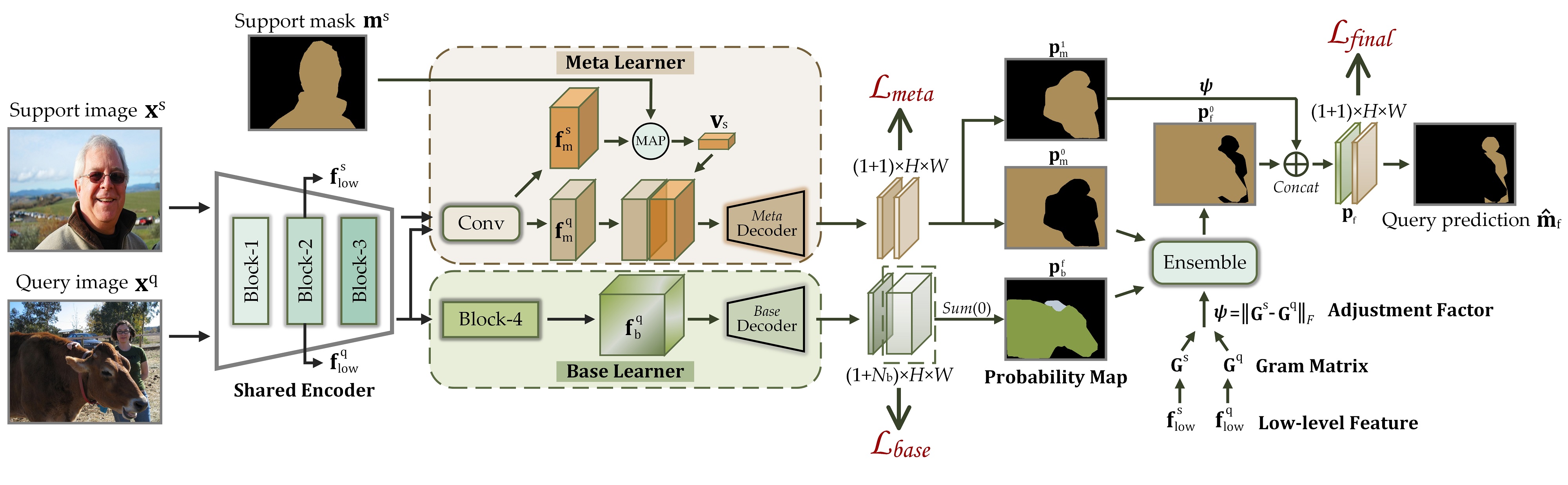}
	\vspace{-0.5em}
	\caption{Overall architecture of the proposed BAM, which is composed of three essential components: a base learner, a meta learner, and an ensemble module. In each training episode, the two learners extract the features of input image pairs $\left( {{{\bf{x}}^{\rm{s}}},{{\bf{x}}^{\rm{q}}}} \right)$ with a shared encoder and make predictions for the specific base category $c$ (note that $c$ denotes the novel category in the meta-testing phase) and the remaining base categories, respectively. Then, the coarse predictions are fed to the ensemble module along with an adjustment factor $\psi$ to suppress the falsely activated regions of base categories, further producing accurate segmentation results. For ease of understanding, we present the probability maps in the form of segmentation masks, but they are actually two-dimensional floating-point matrices, \textit{i.e.}, ${\bf{p}} \in {\left[ {0,1} \right]^{H \times W}}$. MAP represents the masked average pooling operation \cite{zhang2018sg}.}
	\label{fig:2}
\end{figure*}

\section{Proposed Method}
To alleviate the bias problem of current FSS methods, we propose to build an additional network to explicitly predict the regions of base classes in the query images, thereby facilitating the segmentation of novel objects. Without loss of generality, we present the overall architecture of our model under the 1-shot setting (see Fig.\,\ref{fig:2}). The proposed BAM consists of three major components including two complementary learners (\textit{i.e.}, base learner and meta learner) and an ensemble module. The two learners with a shared backbone are used to recognize the base and novel classes, respectively. Then, the ensemble module receives their coarse predictions and an adjustment factor $\psi$ to suppress the falsely activated regions of base classes, further producing accurate segmentation. Moreover, we also propose to learn the fusion weights of different support images under the \textit{K}-shot setting based on $\psi$, aiming to provide better guidance for the query branch. 

\subsection{Base Learner}
\label{sec:4.1}
As mentioned in Sec.\,\ref{sec:2}, current FSS models are biased towards the seen classes, which impedes the recognition of novel concepts. Based on this observation, we propose to introduce an additional branch, \textit{i.e.}, the base learner, to explicitly predict the regions of base classes in the query images. Specifically, given a query image ${{\bf{x}}^{\rm{q}}} \in \mathbb{R}{^{3 \times H \times W}}$, we first apply the encoder network $\mathcal{E}$ and convolutional block to extract its intermediate feature maps ${\bf{f}}_{\rm{b}}^{\rm{q}}$, which can be formulated as: 
\begin{equation}
\setlength{\abovedisplayskip}{4pt}
\setlength{\belowdisplayskip}{4pt}
{\bf{f}}_{\rm{b}}^{\rm{q}} = {\mathcal{F}_{{\rm{conv}}}}\left( {\mathcal{E}\left( {{{\bf{x}}^{\rm{q}}}} \right)} \right) \in \mathbb{R}{^{c \times h \times w}},
\end{equation}
\renewcommand{\thefootnote}{\fnsymbol{footnote}}where ${\mathcal{F}_{{\rm{conv}}}}$ denotes the sequential convolution operations\footnote{Taking ResNet \cite{he2016deep} feature extractor as an example, ${\mathcal{F}_{{\rm{conv}}}}$ is the last convolutional block, namely $block$4.}. $c,h,w$ are the channel dimension, height, and width, respectively, and $h\times$$w$ indicates the minimum resolution among all extracted feature maps.

Then, the decoder network ${\mathcal{D}_{\rm{b}}}$ progressively enlarges the spatial scale of intermediate feature maps ${\bf{f}}_{\rm{b}}^{\rm{q}}$, and finally yields the prediction results, which can be defined as:
\begin{equation}
{{\bf{p}}_{\rm{b}}} = {\rm{softmax}}\left( {{\mathcal{D}_{\rm{b}}}\left( {{\bf{f}}_{\rm{b}}^{\rm{q}}} \right)} \right) \in \mathbb{R}{^{\left( {1 + {N_{\rm{b}}}} \right) \times H \times W}},
\end{equation}
where ${\rm{softmax}}(\cdot)$ operation is conducted along the channel dimension to generate probability maps ${{\bf{p}}_{\rm{b}}}$. ${N_{\rm{b}}}$ represents the number of base categories\footnote{Typically, ${N_{\rm{b}}=15}$ for PASCAL-5$^i$ \cite{shaban2017one} and 60 for COCO-20$^i$ \cite{nguyen2019feature}.}. \\
\indent Unlike the episodic learning paradigm widely adopted in few-shot scenarios, we follow the standard supervised learning paradigm to train the base learner. The cross entropy (CE) loss is leveraged to evaluate the difference between the prediction ${{\bf{p}}_{\rm{b}}}$ and the ground-truth ${\bf{m}}_{\rm{b}}^{\rm{q}}$ at all spatial locations, which can be denoted as: 
\begin{equation}
\setlength{\abovedisplayskip}{4pt}
\setlength{\belowdisplayskip}{4pt}
{\mathcal{L}_{{\rm{base}}}} = \frac{1}{{{n_{{\rm{bs}}}}}}\sum\limits_{i = 1}^{{n_{{\rm{bs}}}}} {{\rm{CE}}\left( {{{\bf{p}}_{{\rm{b;}}i}},{\bf{m}}_{{\rm{b}};i}^{\rm{q}}} \right)},
\end{equation}
where ${n_{{\rm{bs}}}}$ is the number of training samples in each batch.
\vspace{0.1cm}\\
\textbf{Why not train two learners jointly?} A natural way to predict the regions of base classes in the query images is to follow the standard semantic segmentation network, such as PSPNet \cite{zhao2017pyramid}, DeepLab \cite{chen2017deeplab}, etc. However, it is unrealistic to additionally build such a large network on the basis of the original few-shot model, which will introduce too many parameters and slow down the inference speed. Therefore, we attempt to design a unified framework in which two learners share the same backbone network. Nevertheless, we notice that the advanced FSS methods \cite{zhang2019canet,tian2020prior,li2021adaptive} typically freeze the backbone network during training to enhance generalization. Such an operation is inconsistent with the learning scheme of the standard segmentation model and will undoubtedly affect the performance of the base learner. More importantly, it is unknown whether the base learner can be trained well with the episodic learning paradigm, so a two-stage training strategy is eventually adopted. In Sec.\,\ref{sec:5.3}, we will discuss the effects of different training methods and network designs on segmentation accuracy. 

\subsection{Meta Learner}
\label{sec:4.2}
Given a support set $\mathcal{S}=\left\{ {{{\bf{x}}^{\rm{s}}},{{\bf{m}}^{\rm{s}}}} \right\}$ and a query image ${{\bf{x}}^{\rm{q}}}$, the goal of the meta learner is to segment the objects in ${{\bf{x}}^{\rm{q}}}$ that share the same category as the annotation mask ${{\bf{m}}^{\rm{s}}}$ under the guidance of $\mathcal{S}$. In our work, we first follow \cite{zhang2019canet,tian2020prior} to concatenate the features derived from $block$2 and $block$3. Then, a $1\times$$1$ convolution is applied to reduce the channel dimension and generate intermediate feature maps:
\begin{equation}
{\bf{f}}_{\rm{m}}^{\rm{s}} = {\mathcal{F}_{{\rm{1}} \times {\rm{1}}}}\left( {\mathcal{E}\left( {{{\bf{x}}^{\rm{s}}}} \right)} \right) \in \mathbb{R}{^{c \times h \times w}},
\end{equation}
\begin{equation}
{\bf{f}}_{\rm{m}}^{\rm{q}} = {\mathcal{F}_{{\rm{1}} \times {\rm{1}}}}\left( {\mathcal{E}\left( {{{\bf{x}}^{\rm{q}}}} \right)} \right) \in \mathbb{R}{^{c \times h \times w}},
\end{equation}
where $\mathcal{E}$ is the encoder network shared with both base and meta learners, and ${\mathcal{F}_{{\rm{1}} \times {\rm{1}}}}$ denotes the $1\times$$1$ convolution that encodes the input features to 256 dimensions. Furthermore, we calculate the prototype through the masked average pooling (MAP) \cite{zhang2018sg} w.r.t. $\left( {{\bf{f}}_{\rm{m}}^{\rm{s}},{{\bf{m}}^{\rm{s}}}} \right)$ to provide crucial class-related cues: 
\begin{equation}
{{\bf{v}}_{\rm{s}}} = {\mathcal{F}_{{\rm{pool}}}}\left( {{\bf{f}}_{\rm{m}}^{\rm{s}} \odot \mathcal{I}\left( {{{\bf{m}}^{\rm{s}}}} \right)} \right) \in \mathbb{R}{^c},
\end{equation}
where $\mathcal{F}_{{\rm{pool}}}$ is the average-pooling operation, $ \odot $ represents Hadamard product, and $\mathcal{I}$ is a function that reshapes ${{\bf{m}}^{\rm{s}}}$ to be the same shape as ${\bf{f}}_{\rm{m}}^{\rm{s}}$ through interpolation and expansion techniques such that $\mathcal{I}:{\mathbb{R}^{H \times W}} \to {\mathbb{R}^{c \times h \times w}}$. Afterwards, the target regions in ${\bf{f}}_{\rm{m}}^{\rm{q}}$ are activated under the guidance of ${{\bf{v}}_{\rm{s}}}$, and the final prediction results are generated through the decoder network, which can be summarized as:
\begin{equation}
{{\bf{p}}_{\rm{m}}} = {\rm{softmax}}\left( {{\mathcal{D}_{\rm{m}}}\left( {{\mathcal{F}_{{\rm{guidance}}}}\left( {{{\bf{v}}_{\rm{s}}},{\bf{f}}_{\rm{m}}^{\rm{q}}} \right)} \right)} \right) \in \mathbb{R}{^{2 \times H \times W}},
\end{equation}
where $\mathcal{D}_{\rm{m}}$ denotes the decoder network of the meta learner. ${\mathcal{F}_{{\rm{guidance}}}}$ is an essential module of FSS that passes the annotation information from the support branch to the query branch to provide specific segmentation cues. It represents the ``expand \& concatenate'' operations \cite{zhang2019canet} in our work. Similarly, we calculate the BCE loss between ${{\bf{p}}_{\rm{m}}}$ and ${{\bf{m}}^{\rm{q}}}$ to update all parameters of the meta learner: 
\begin{equation}
\label{eq:8}
\setlength{\abovedisplayskip}{4pt}
\setlength{\belowdisplayskip}{4pt}
{\mathcal{L}_{{\rm{meta}}}} = \frac{1}{{{n_{\rm{e}}}}}\sum\limits_{i = 1}^{{n_{\rm{e}}}} {{\rm{BCE}}\left( {{{\bf{p}}_{\rm{m}}}_{{\rm{;}}i},{\bf{m}}_i^{\rm{q}}} \right)},
\end{equation}
where ${n_{\rm{e}}}$ denotes the number of training episodes in each batch. 

\subsection{Ensemble}
\label{sec:4.3}
Considering that meta learners are typically sensitive to the quality of support images, we further propose to leverage the evaluation results of the scene differences between query-support image pairs to adjust the coarse predictions derived from meta learners. Specifically, we first integrate the foreground probability maps generated by the base learner to obtain the prediction of the background region relative to the few-shot task: 
\begin{equation}
\setlength{\abovedisplayskip}{4pt}
\setlength{\belowdisplayskip}{4pt}
{\bf{p}}_{\rm{b}}^{\rm{f}} = \sum\limits_{i = 1}^{{N_{\rm{b}}}} {{\bf{p}}_{\rm{b}}^i},
\end{equation}
where the superscript of ${\bf{p}}_{\rm{b}}^{\rm{f}}$ stands for the foreground, and the subscript ``b'' stands for the base learner. 

\begin{figure}[t]
	\centering
	\includegraphics[width=1.0\linewidth]{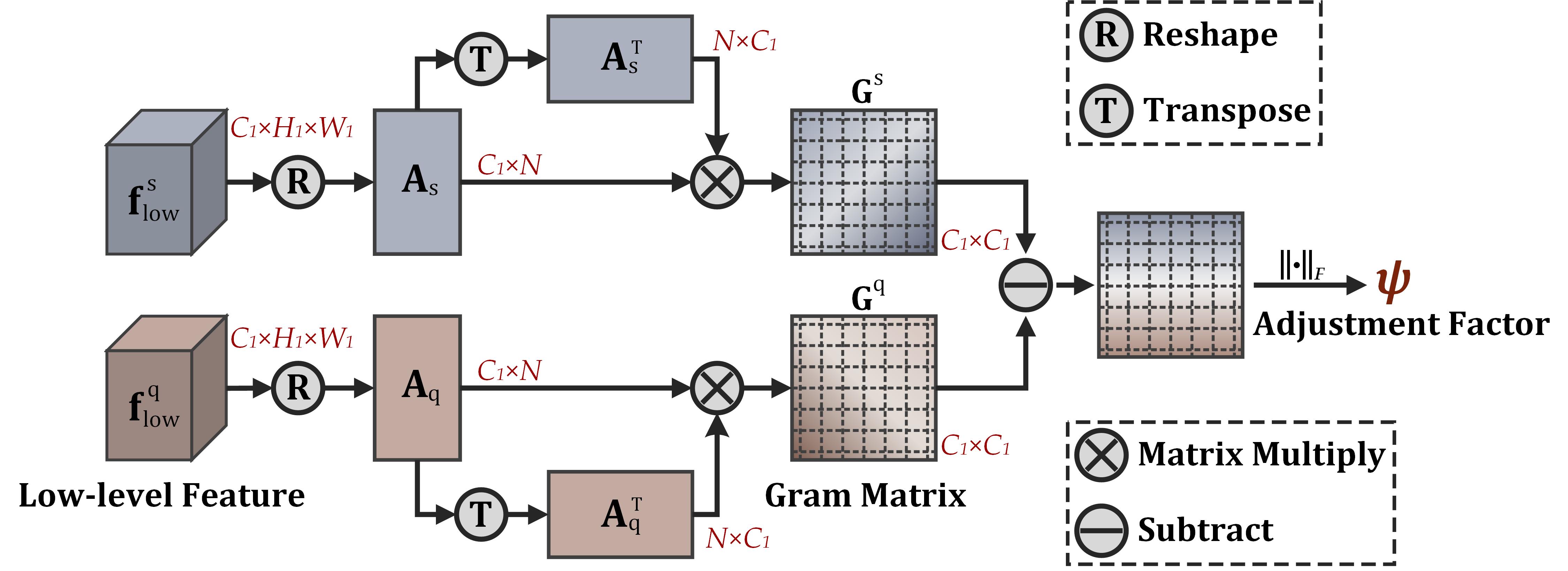}
	\caption{The calculation process of the adjustment factor $\psi$ for the low-level features ${\bf{f}}_{{\rm{low}}}^{\rm{s}}$ and ${\bf{f}}_{{\rm{low}}}^{\rm{q}}$.}
	\label{fig:3}
\end{figure}

Then, we leverage the low-level features ${\bf{f}}_{{\rm{low}}}^{\rm{s}},{\bf{f}}_{{\rm{low}}}^{\rm{q}} \in \mathbb{R}{^{{C_1} \times {H_1} \times {W_1}}}$ extracted from the fixed backbone network to calculate the Gram matrices of support and query images, respectively (see Fig.\,\ref{fig:3}). Please note that the relevant operations of these two input images are similar, and that of the support one can be summarized as: 
\begin{equation}
\setlength{\abovedisplayskip}{7pt}
\setlength{\belowdisplayskip}{7pt}
{{\bf{A}}_{\rm{s}}} = {\mathcal{F}_{{\rm{reshape}}}}\left( {{\bf{f}}_{{\rm{low}}}^{\rm{s}}} \right) \in \mathbb{R}{^{{C_1} \times N}},
\end{equation}
\begin{equation}
{{\bf{G}}^{\rm{s}}}{\rm{ = }}{{\bf{A}}_{\rm{s}}}{\bf{A}}_{\rm{s}}^\mathsf{T} \in \mathbb{R}{^{{C_1} \times {C_1}}},
\end{equation}
where $N$$=$${H_1}\times$${W_1}$ and ${\mathcal{F}_{{\rm{reshape}}}}$ reshapes the size of the input tensor to ${C_1}\times$$N$. With the calculated Gram matrices  , the Frobenius norm is evaluated on their difference to obtain the overall indicator $\psi$ for guiding the adjustment process: 
\begin{equation}
\psi  = {{\Vert{{\bf{G}}^{\rm{s}}} - {{\bf{G}}^{\rm{q}}}\Vert}_F},
\end{equation}
where ${\left\| {\,\cdot\,} \right\|_F}$ denotes the Frobenius norm of the input matrix. After that, the coarse results of the two learners are integrated under the guidance of adjustment factor $\psi$, further yielding the final segmentation predictions ${{\bf{p}}_{\rm{f}}}$:
\begin{equation}
\setlength{\abovedisplayskip}{7pt}
\setlength{\belowdisplayskip}{7pt}
{\bf{p}}_{\rm{f}}^{\rm{0}} = {\mathcal{F}_{{\rm{ensemble}}}}\left( {{\mathcal{F}_\psi }\left( {{\bf{p}}_{\rm{m}}^0} \right),{\bf{p}}_{\rm{b}}^{\rm{f}}} \right),
\end{equation}
\begin{equation}
{{\bf{p}}_{\rm{f}}}{\rm{ = }}{\bf{p}}_{\rm{f}}^{\rm{0}} \oplus {\mathcal{F}_\psi }\left( {{\bf{p}}_{\rm{m}}^1} \right),
\end{equation}
where ${{\bf{p}}_{\rm{m}}},{{\bf{p}}_{\rm{b}}}$  denote the predictions of the meta learner and base learner respectively. The superscript ``0'' and ``1'' represent the background and foreground respectively. Both ${\mathcal{F}_\psi }$ and ${\mathcal{F}_{{\rm{ensemble}}}}$ are $1\times$$1$ convolution operations with specific initial parameters. The goal of the former is to adjust the coarse results of the meta learner, while the goal of the latter is to integrate the two learners. $\oplus$ indicates the concatenation operation along channel dimension. Finally, the overall loss during the meta-training phase can be evaluated by: 
\begin{equation}
\mathcal{L} = {\mathcal{L}_{{\rm{final}}}} + \lambda {\mathcal{L}_{{\rm{meta}}}},
\end{equation}
\begin{equation}
{\mathcal{L}_{{\rm{final}}}} = \frac{1}{{{n_{\rm{e}}}}}\sum\limits_{i = 1}^{{n_{\rm{e}}}} {{\rm{BCE}}\left( {{\bf{p}}_i^{\rm{q}},{\bf{m}}_i^{\rm{q}}} \right)},
\end{equation}
where $\lambda$ is set to $1.0$ in all experiments, and ${\mathcal{L}_{{\rm{meta}}}}$ is the loss function of the meta learner defined by Eq.\,(\ref{eq:8}). 

\subsection{K-Shot Setting}
\label{sec:4.4}
When the task is extended to the $K$-shot $(K$$>$$1)$ setting, more than one annotated (support) images are available. Current FSS methods typically average the \textit{prototypes} extracted from the support branch and then utilize the averaged features to guide the subsequent segmentation process, which assumes that the contribution of each sample is the same \cite{wang2019panet,tian2020prior}. However, such an approach might be suboptimal since the samples with significant scene differences from the query images cannot provide more-targeted guidance. Therefore, we further propose to adaptively estimate the weight of each support image based on the adjustment factor $\psi$, where a smaller value indicates a greater contribution and vice versa. 

Specifically, given the adjustment factor ${\psi _i}$ of each support sample, we first integrate them into a unified vector ${\psi _{\rm{t}}} \in \mathbb{R}{^K}$ through the concatenation operation. Then, two fully connected (FC) layers are applied to generate the fusion weights $\eta$ of the support images: 
\begin{equation}
\eta  = {\rm{soft}}\max \left( {{\bf{w}}_2^\mathsf{T}{\rm{ReLU}}\left( {{\bf{w}}_1^\mathsf{T}{\psi _{\rm{t}}}} \right)} \right) \in \mathbb{R}{^K},
\end{equation}
where ${{\bf{w}}_1} \in \mathbb{R}{^{K \times \frac{K}{r}}},{{\bf{w}}_2} \in \mathbb{R}{^{\frac{K}{r} \times K}}$ are the weights of the FC layers, and $r$ represents the dimensionality reduction factor. At last, we make a weighted summation to achieve the final $\psi$ for ensemble. 

\begin{table*}[t]
	\centering
	\resizebox{0.75\linewidth}{!}{
		\renewcommand{\arraystretch}{1.1}
		\begin{tabular}{p{1.3cm}>{\hfill}p{3.7cm}|p{1.0cm}<{\centering}p{1.0cm}<{\centering}p{1.0cm}<{\centering}p{1.0cm}<{\centering}p{1.0cm}<{\centering}|p{1.0cm}<{\centering}p{1.0cm}<{\centering}p{1.0cm}<{\centering}p{1.0cm}<{\centering}p{1.0cm}<{\centering}}
			\hline
			\multirow{2}{*}{Backbone} & \multirow{2}{*}{Method} & \multicolumn{5}{c|}{1-shot}           & \multicolumn{5}{c}{5-shot}            \\ \cline{3-12} 
			&            & Fold-0 & Fold-1 & Fold-2 & Fold-3 & Mean  & Fold-0 & Fold-1 & Fold-2 & Fold-3 & Mean  \\ \hline
			\multirow{8}{*}{VGG16}    & SG-One (TCYB'19)\cite{zhang2018sg}                 & 40.20 & 58.40 & 48.40 & 38.40 & 46.30 & 41.90 & 58.60 & 48.60 & 39.40 & 47.10 \\
			& PANet (ICCV'19)\cite{wang2019panet}     & 42.30  & 58.00  & 51.10  & 41.20  & 48.10 & 51.80  & 64.60  & 59.80  & 46.50  & 55.70 \\
			& FWB (ICCV'19)\cite{wang2019panet}       & 47.00  & 59.60  & 52.60  & 48.30  & 51.90 & 50.90  & 62.90  & 56.50  & 50.10  & 55.10 \\
			& CRNet (CVPR'20)\cite{liu2020crnet}     & -      & -      & -      & -      & 55.20 & -      & -      & -      & -      & 58.50 \\
			& PFENet (TPAMI'20)\cite{tian2020prior}    & 56.90  & \underline{68.20}  & 54.40  & 52.40  & 58.00 & 59.00  & 69.10  & 54.80  & 52.90  & 59.00 \\
			& HSNet (ICCV'21)\cite{min2021hypercorrelation}     & 59.60  & 65.70  & 59.60  & 54.00  & 59.70 & \underline{64.90}  & 69.00  & 64.10  & 58.60  & 64.10 \\ \cline{2-12} 
			& Baseline   & \underline{59.90}  & 67.51  & \underline{64.93}  & \underline{55.72}  & \underline{62.0}2 & 64.02  & \underline{71.51}  & \underline{69.39}  & \underline{63.55}  & \underline{67.12} \\
			& \cellcolor{mygray}BAM (ours) & \cellcolor{mygray}\textbf{63.18}  & \cellcolor{mygray}\textbf{70.77}  & \cellcolor{mygray}\textbf{66.14}  & \cellcolor{mygray}\textbf{57.53}  & \cellcolor{mygray}\textbf{64.41} & \cellcolor{mygray}\textbf{67.36}  & \cellcolor{mygray}\textbf{73.05}  & \cellcolor{mygray}\textbf{70.61}  & \cellcolor{mygray}\textbf{64.00}  & \cellcolor{mygray}\textbf{68.76} \\ \hline
			\multirow{8}{*}{ResNet50} & CANet (ICCV'19)\cite{zhang2019canet}                  & 52.50 & 65.90 & 51.30 & 51.90 & 55.40 & 55.50 & 67.80 & 51.90 & 53.20 & 57.10 \\
			& PGNet (ICCV'19)\cite{zhang2019pyramid}     & 56.00  & 66.90  & 50.60  & 50.40  & 56.00 & 57.70  & 68.70  & 52.90  & 54.60  & 58.50 \\
			& CRNet (CVPR'20)\cite{liu2020crnet}     & -      & -      & -      & -      & 55.70 & -      & -      & -      & -      & 58.80 \\			
			& PPNet (ECCV'20)\cite{liu2020part}     & 48.58  & 60.58  & 55.71  & 46.47  & 52.84 & 58.85  & 68.28  & 66.77  & 57.98  & 62.97 \\
			& PFENet (TPAMI'20)\cite{tian2020prior}    & 61.70  & 69.50  & 55.40  & 56.30  & 60.80 & 63.10  & 70.70  & 55.80  & 57.90  & 61.90 \\
			& HSNet (ICCV'21)\cite{min2021hypercorrelation}     & 64.30  & 70.70  & 60.30  &  \underline{60.50}  & 64.00 &  \underline{70.30}  &  \underline{73.20}  & 67.40  &  \underline{67.10}  &  \underline{69.50} \\ \cline{2-12} 
			& Baseline   &  \underline{65.68}  &  \underline{71.41}  &  \underline{65.56}  & 58.93  &  \underline{65.40} & 67.28  & 72.38  &  \underline{69.16}  & 66.25  & 68.77 \\
			& \cellcolor{mygray}BAM (ours) & \cellcolor{mygray}\textbf{68.97}  & \cellcolor{mygray}\textbf{73.59}  & \cellcolor{mygray}\textbf{67.55}  & \cellcolor{mygray}\textbf{61.13}  & \cellcolor{mygray}\textbf{67.81} & \cellcolor{mygray}\textbf{70.59}  & \cellcolor{mygray}\textbf{75.05}  & \cellcolor{mygray}\textbf{70.79}  & \cellcolor{mygray}\textbf{67.20}  & \cellcolor{mygray}\textbf{70.91} \\ \hline
	\end{tabular}}
	\caption{Performance comparison on PASCAL-5$^i$ in terms of mIoU. ``Baseline'' means the meta learner that shares the encoder network $\mathcal{E}$ \texttt{pre-trained} by the base learner. Results in \textbf{bold} denote the best performance, while the \underline{underlined} ones indicate the second best.}
	\label{tab:1}
\end{table*}

\begin{table*}[t]
	\vspace{0.3cm}
	\begin{minipage}[t]{0.7\textwidth}
		\centering
		\resizebox{0.95\linewidth}{1.8cm}{
			\renewcommand{\arraystretch}{1.05}
			\begin{tabular}{ll|ccccc|ccccc}
				\hline
				\multirow{2}{*}{Backbone} & \multirow{2}{*}{Method} & \multicolumn{5}{c|}{1-shot}           & \multicolumn{5}{c}{5-shot}            \\ \cline{3-12} 
				&            & Fold-0 & Fold-1 & Fold-2 & Fold-3 & Mean  & Fold-0 & Fold-1 & Fold-2 & Fold-3 & Mean  \\ \hline
				\multirow{5}{*}{VGG16}    & FWB \cite{nguyen2019feature}                    & 18.35 & 16.72 & 19.59 & 25.43 & 20.02 & 20.94 & 19.24 & 21.94 & 28.39 & 22.63 \\
				& PFENet \cite{tian2020prior}    & 35.40  & 38.10  & 36.80  & 34.70  & 36.30 & 38.20  & 42.50  & 41.80  & 38.90  & 40.40 \\
				& PRNet \cite{liu2020prototype}     & 27.46  & 32.99  & 26.70  & 28.98  & 29.03 & 31.18  & 36.54  & 31.54  & 32.00  & 32.82 \\ \cline{2-12} 
				& Baseline   & \underline{38.42}  & \underline{43.75}  & \underline{44.32}  & \underline{39.84}  & \underline{41.58} & \underline{45.93}  & \underline{48.88}  & \underline{47.87}  & \underline{46.96}  & \underline{47.41} \\
				& \cellcolor{mygray}BAM (ours) & \cellcolor{mygray}\textbf{38.96}  & \cellcolor{mygray}\textbf{47.04}  & \cellcolor{mygray}\textbf{46.41}  & \cellcolor{mygray}\textbf{41.57}  & \cellcolor{mygray}\textbf{43.50} & \cellcolor{mygray}\textbf{47.02}  & \cellcolor{mygray}\textbf{52.62}  & \cellcolor{mygray}\textbf{48.59}  & \cellcolor{mygray}\textbf{49.11}  & \cellcolor{mygray}\textbf{49.34} \\ \hline
				\multirow{5}{*}{ResNet50} & HFA \cite{liu2021harmonic}       & 28.65  & 36.02  & 30.16  & 33.28  & 32.03 & 32.69  & 42.12  & 30.35  & 36.19  & 35.34 \\
				& ASGNet \cite{li2021adaptive}    & -      & -      & -      & -      & 34.56 & -      & -      & -      & -      & 42.48 \\
				& HSNet \cite{min2021hypercorrelation}     & 36.30  & 43.10  & 38.70  & 38.70  & 39.20 & 43.30  & 51.30  & 48.20  & 45.00  & 46.90 \\ \cline{2-12} 
				& Baseline   & \underline{41.92}  & \underline{45.35}  & \underline{43.86}  & \underline{41.24}  & \underline{43.09} & \underline{46.98}  & \underline{51.87}  & \underline{49.49}  & \underline{47.81}  & \underline{49.04} \\
				& \cellcolor{mygray}BAM (ours) & \cellcolor{mygray}\textbf{43.41}  & \cellcolor{mygray}\textbf{50.59}  & \cellcolor{mygray}\textbf{47.49}  & \cellcolor{mygray}\textbf{43.42}  & \cellcolor{mygray}\textbf{46.23} & \cellcolor{mygray}\textbf{49.26}  & \cellcolor{mygray}\textbf{54.20}  & \cellcolor{mygray}\textbf{51.63}  & \cellcolor{mygray}\textbf{49.55}  & \cellcolor{mygray}\textbf{51.16} \\ \hline
		\end{tabular}}
		\caption{Performance comparison on COCO-20$^i$ in terms of mIoU. ``Baseline'' means the meta learner with \texttt{pre-trained} $\mathcal{E}$ in our work.}
		\label{tab:2}
	\end{minipage}
	\hspace{0.3cm}
	\begin{minipage}[t]{0.25\textwidth}
		\centering
		\resizebox{0.95\linewidth}{1.3cm}{
			\renewcommand{\arraystretch}{1.05}
			\begin{tabular}{p{1.5cm}p{2cm}|p{1.5cm}<{\centering}p{1.5cm}<{\centering}}
				\hline
				\multirow{2}{*}{Backbone} & \multirow{2}{*}{Method} & \multicolumn{2}{c}{FB-IoU (\%)} \\ \cline{3-4} 
				&                         & 1-shot         & 5-shot         \\ \hline
				\multirow{5}{*}{VGG16}    & OSLSM \cite{shaban2017one}                  & 61.30          & 61.50          \\
				& co-FCN \cite{rakelly2018conditional}                 & 60.10          & 60.20          \\
				& PFENet \cite{tian2020prior}                 & 72.00          & 72.30          \\
				& HSNet \cite{min2021hypercorrelation}                  & \underline{73.40}          & \underline{76.60}          \\
				& \cellcolor{mygray}BAM (ours)              & \cellcolor{mygray}\textbf{77.26}          & \cellcolor{mygray}\textbf{81.10}          \\ \hline
				\multirow{5}{*}{ResNet50} & PGNet \cite{zhang2019pyramid}                  & 69.90          & 70.50          \\
				& PPNet \cite{liu2020part}                  & 69.19          & 75.76          \\
				& PFENet \cite{tian2020prior}                 & 73.30          & 73.90          \\
				& HSNet \cite{min2021hypercorrelation}                  & \underline{76.70}          & \underline{80.60}          \\
				& \cellcolor{mygray}BAM (ours)              & \cellcolor{mygray}\textbf{79.71}          & \cellcolor{mygray}\textbf{82.18}          \\ \hline
		\end{tabular}}
		\caption{Averaged FB-IoU over 4 folds on PASCAL-5$^i$.}
		\label{tab:3}
	\end{minipage}
\end{table*}

\subsection{Extension to Generalized FSS}
\label{sec:4.5}
The proposed BAM is originally designed for standard FSS tasks, but it could be easily extended to generalized settings, where the regions of base and novel classes in the query images are required to be determined. In this work, we simply fuse the results of the base learner and the final results after ensemble according to a predefined threshold $\tau$ to obtain the holistic segmentation predictions ${{\bf{\hat m}}_{\rm{g}}}$, which can be formulated as:
\begin{equation}
\label{eq:18}
{\bf{\hat m}}_{\rm{g}}^{(x,y)}{\rm{ = }}\left\{ \begin{array}{l}
1{\kern 2.8em}{\bf{p}}_{\rm{f}}^{1;\left( {x,y} \right)} > \tau \\
{\bf{\hat m}}_{\rm{b}}^{(x,y)}{\kern 1.2ex}{\bf{p}}_{\rm{f}}^{1;\left( {x,y} \right)} \le \tau \;{\rm{and }}\;{\bf{\hat m}}_{\rm{b}}^{(x,y)} \ne {\rm{0 }}\\
0{\kern 2.8em}{\rm{otherwise}}
\end{array} \right.,
\end{equation}
where $\left( {x,y} \right)$ denotes the spatial location. ${{\bf{\hat m}}_{\rm{b}}}$ represents the base segmentation masks, which can be calculated by: 
\begin{equation}
\setlength{\abovedisplayskip}{4pt}
\setlength{\belowdisplayskip}{4pt}
{{\bf{\hat m}}_{\rm{b}}}{\rm{ = }}\arg \max \left( {{{\bf{p}}_{\rm{b}}}} \right) \in {\left\{ {0,1,...,{N_{\rm{b}}}} \right\}^{H \times W}},
\end{equation}
where $\arg \max (\cdot)$ is performed along the channel dimension. 

\begin{figure*}[t]
	\centering
	\includegraphics[width=0.95\linewidth]{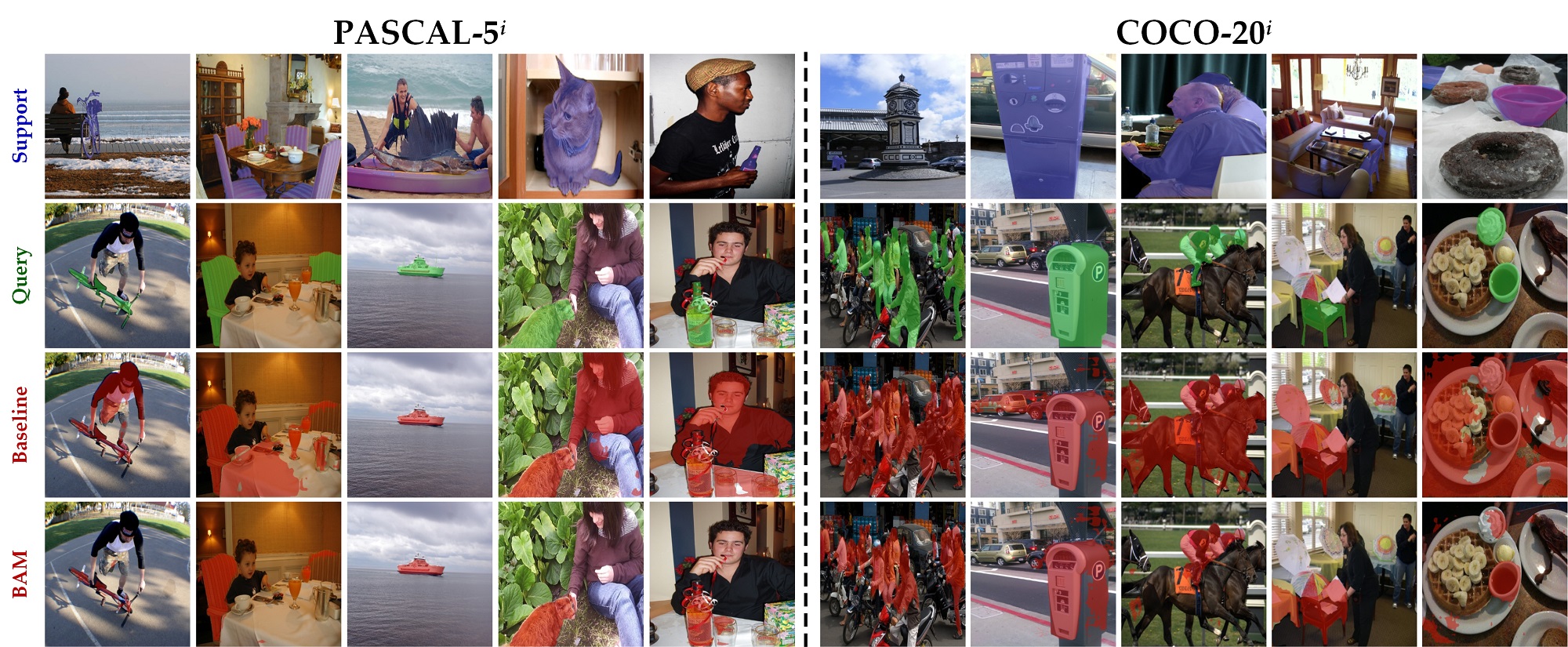}
	\caption{Qualitative results of the proposed BAM and baseline approach under 1-shot setting. The left panel is from PASCAL-5$^i$, and the right one is from COCO-20$^i$. Each row from top to bottom represents the support images with ground-truth (GT) masks (\textcolor[rgb]{0,0,1}{blue}), query images with GT masks (\textcolor[rgb]{0,1,0}{green}), baseline results (\textcolor[rgb]{1,0,0}{red}), and our results (\textcolor[rgb]{1,0,0}{red}), respectively.}
	\label{fig:4}
\end{figure*}

\section{Experiments}
\subsection{Setup}
\label{sec:5.1}
\noindent\textbf{Datasets.} We evaluate the performance of our approach on two widely-used FSS datasets, namely PASCAL-5$^i$ \cite{shaban2017one} and COCO-20$^i$ \cite{nguyen2019feature}. PASCAL-5$^i$ is proposed by Shaban \textit{et al}. and created from PASCAL VOC 2012 \cite{everingham2010pascal} with additional annotations from SDS \cite{hariharan2011semantic}, while COCO-20$^i$ is presented in \cite{nguyen2019feature} and built from MSCOCO \cite{lin2014microsoft}. The object categories of both datasets are evenly divided into four folds, and the experiments are conducted in a cross-validation manner. For each fold, we randomly sample 1,000 pairs of support and query images for validation. \vspace{0.05cm}\\
\textbf{Evaluation metrics.} Following the previous works \cite{yang2020prototype,tian2020prior,liu2020part}, we adopt mean intersection-over-union (mIoU) and foreground-background IoU (FB-IoU) as the evaluation metrics for experiments.\vspace{0.05cm}\\
\textbf{Implementation details.} The training process of the proposed approach can be divided into two stages, \textit{i.e.}, pre-training and meta-training. For the first stage, we adopt the standard supervised learning paradigm to train the base learner on each fold of the FSS dataset, which consists of 16/61 classes (including background) for PASCAL-5$^i$/COCO-20$^i$. PSPNet \cite{zhao2017pyramid} is served as the base learner in our work, and it is trained on PASCAL-5$^i$ for 100 epochs and COCO-20$^i$ for 20 epochs. SGD optimizer with initial learning rate 2.5e-3 is used for updating the parameters, and the training batch size is set to 12. For the second stage, we jointly train the meta learner and ensemble module in an episodic learning fashion, and the parameters of the base learner are fixed in this stage. Note that two learners share the same encoder to extract the features of input images, which is also not optimized to facilitate generalization. The rest of the network layers are trained with SGD optimizer on PASCAL-5$^i$ for 200 epochs and COCO-20$^i$ for 50 epochs. The batch size and learning rate are set to 8 and 5e-2 respectively on both datasets. We follow the data augmentation techniques in \cite{tian2020prior} for training. A variant of PFENet \cite{tian2020prior} is served as the meta learner in our work, where the FEM is replaced by ASPP \cite{chen2017deeplab} to reduce complexity. We average the results of 5 trails with different random seeds. The proposed model is implemented in PyTorch and runs on NVIDIA RTX 2080Ti GPUs. 

\subsection{Comparison with State-of-the-Arts}
\label{sec:5.2}
\noindent\textbf{Quantitative results.} Tables \ref{tab:1} and \ref{tab:2} present the mIoU results of different approaches on PASCAL-5$^i$ and COCO-20$^i$ benchmarks. It can be found that our BAM outperforms the advanced FSS models with a considerable margin and sets new state-of-the-arts under all settings. With VGG16 backbone, the proposed method achieves 4.71\%p (1-shot) and 4.66\%p (5-shot) of mIoU improvements over previous best results on PASCAL-5$^i$. As for COCO-20$^i$, our 1-shot and 5-shot results respectively surpass the best competitor, \textit{i.e.}, HSNet, by 7.03\%p and 4.26\%p mIoU with ResNet50 backbone, demonstrating its remarkable capability of handling complex tasks. Moreover, we also make comparisons of our model with other advanced approaches in terms of FB-IoU on PASCAL-5$^i$ (see Tab.\,\ref{tab:3}). Once again, the proposed BAM achieves substantial improvements, especially for the 1-shot results with ResNet50 backbone.
\vspace{0.05cm}\\
\textbf{Qualitative results.} To better analyze and understand the proposed model, we further take several episodes during the meta-testing phase and visualize the corresponding segmentation results, as shown in Fig.\,\ref{fig:4}. It can be found in our results (4$^{\rm th}$\;row), the falsely activated targets of base classes are significantly suppressed compared to the baseline method (3$^{\rm rd}$\;row), which verify the effectiveness of the base learner and ensemble module. 
\begin{figure}[t]
	\vspace{0.3cm}
	\setlength{\abovecaptionskip}{0pt}
	\centering
	\includegraphics[width=1.0\linewidth]{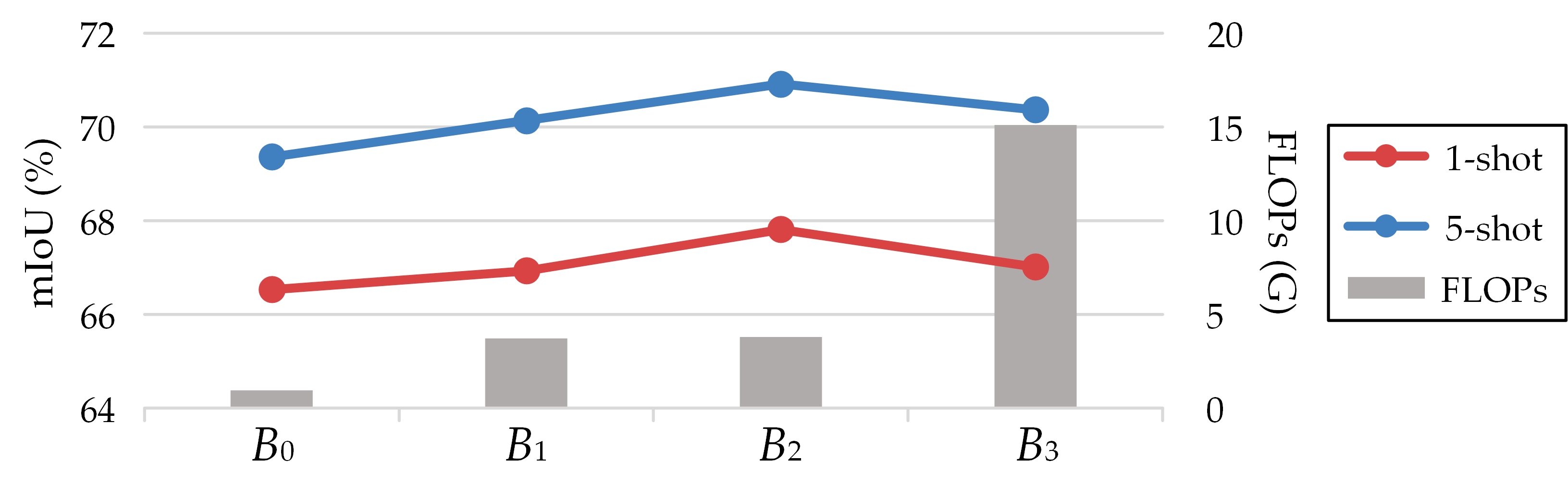}
	\caption{Ablation studies on the low-level features ${{\bf{f}}_{{\rm{low}}}}$ with ResNet50 backbone. ${B_i}$ denotes the feature maps extracted from the $i$-th convolutional blocks of backbone network. FLOPs means floating point operations per second.}
	\label{fig:5}
\end{figure}
\begin{table}[t]	
	\centering
	\renewcommand{\arraystretch}{1.05}
	\resizebox{0.8\linewidth}{!}{
		\begin{tabular}{p{1.3cm}<{\centering}p{1.3cm}<{\centering}p{1.3cm}<{\centering}p{1.3cm}<{\centering}p{1.3cm}<{\centering}p{1.3cm}<{\centering}}
			\hline
			PT & ${\mathcal{L}_{{\rm{meta}}}}$ & Init. & $\bm \psi$ & mIoU & FB-IoU \\ \hline
			&  &  &  & 57.61 & 70.75 \\
			\ding{51} &  &  &  & 59.12 & 71.94 \\
			\ding{51}& \ding{51} &  &  & 59.76 & 72.79 \\
			\ding{51}& \ding{51} & \ding{51} &  & \underline{62.49} & \underline{75.43} \\
			\ding{51}& \ding{51} & \ding{51} & \ding{51} & \textbf{64.41} & \textbf{77.26} \\ \hline
	\end{tabular}}
	\caption{Ablation studies of different design choices under the 1-shot setting. ``PT'' denotes the pre-training for base learner. ``Init.'' represents the specific initial weights of the ensemble module.}
	\label{tab:4}
\end{table}
\subsection{Ablation Study}
\label{sec:5.3}
We conduct a series of ablation studies to investigate the impact of each component on segmentation performance. Note that the experiments in this section are performed on PASCAL-5$^i$ dataset using VGG16 backbone unless specified otherwise. 
\begin{table}[t]
	\vspace{0.4cm}
	\renewcommand{\arraystretch}{1.05}
	\begin{minipage}[t]{0.55\hsize}	
		\centering
		\resizebox{1.0\hsize}{!}{
			\begin{tabular}{lcc}
				\hline
				Method & mIoU (\%) & $\Delta$ \\ \hline
				1-shot baseline & 64.41 & 0 \\ \hline
				Mask-OR \cite{shaban2017one} & 65.15 & 0.74 \\
				Mask-Avg \cite{zhang2019canet} & 65.92 & 1.51 \\
				Feature-Avg \cite{rakelly2018conditional} & \underline{66.83} & \underline{2.42} \\ \hline
				Reweighting (ours) & \textbf{68.76} & \textbf{4.35} \\ \hline
		\end{tabular}}
		\caption{Ablation studies on the 5-shot fusion scheme.}
		\label{tab:5}
	\end{minipage}
	\hspace{0.2cm}
	\begin{minipage}[t]{0.40\hsize}	
		\centering
		\resizebox{1.0\hsize}{!}{
			\begin{tabular}{lcc}
				\hline
				\multirow{2}{*}{Annotation} & \multicolumn{2}{c}{mIoU (\%)} \\ \cline{2-3} 
				& 1-shot & 5-shot \\ \hline
				Pixel-wise labels & 64.41 & 68.76 \\
				Bounding boxes & 62.25 & 66.17 \\ \hline
		\end{tabular}}
		\caption{Ablation studies on support annotations.}
		\label{tab:6}	
	\end{minipage}	
\end{table}
\vspace{-3.2pt}
\begin{figure}[t]
	\centering
	\includegraphics[width=0.89\linewidth]{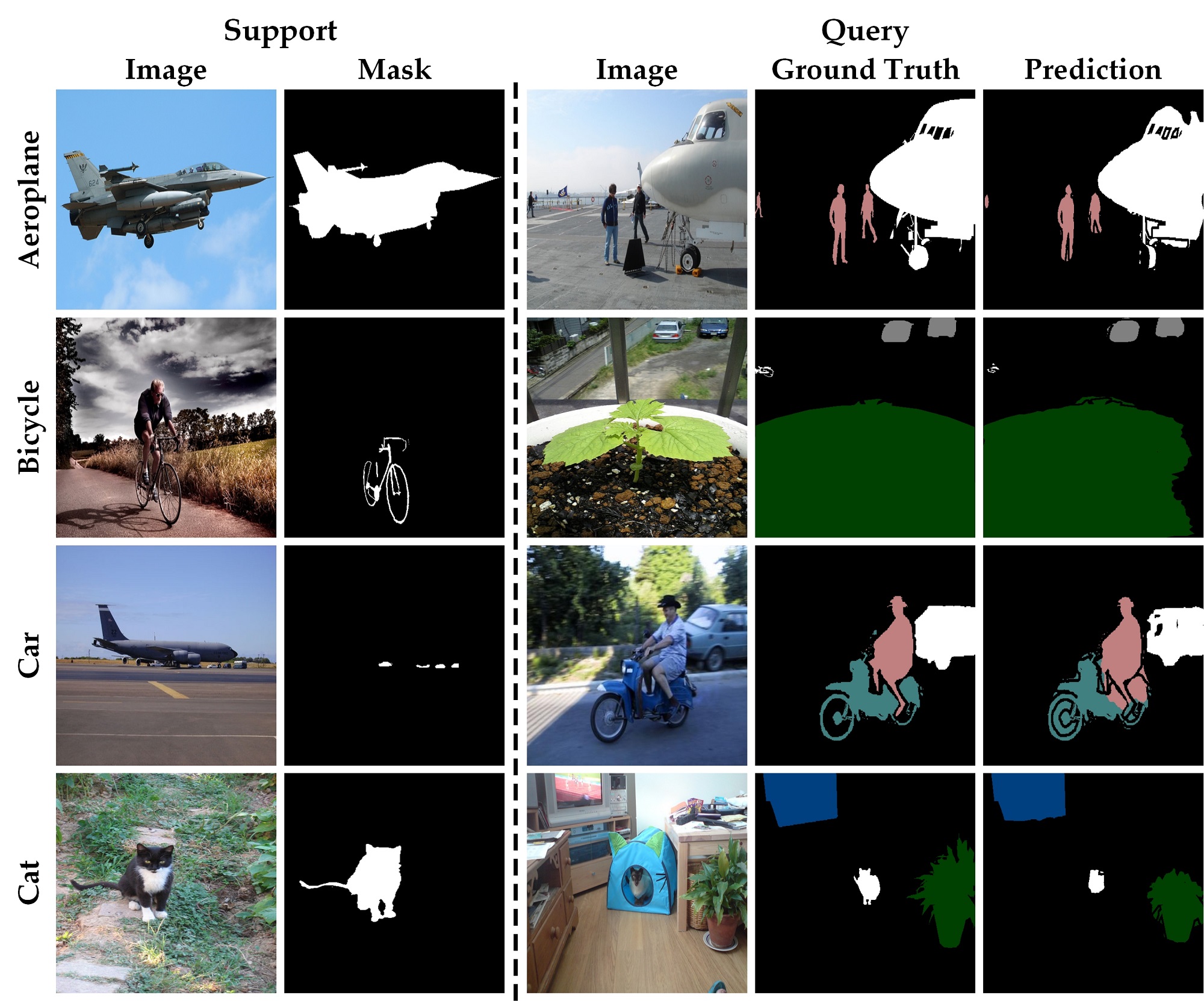}
	\caption{Segmentation results of the proposed approach under generalized FSS setting. Note that white in the query mask represents the novel category, while the other colors represent the base categories. Best viewed in color and zoom in, especially the \textit{bicycle} in the 2$^{\rm nd}$\;row.}
	\label{fig:6}
\end{figure}
\begin{table*}[t]
	\centering
	\renewcommand{\arraystretch}{1.05}
	\resizebox{0.75\linewidth}{!}{
	\begin{tabular}{p{2cm}p{2.5cm}p{1.3cm}<{\centering}p{1.3cm}<{\centering}p{1.3cm}<{\centering}p{1.3cm}<{\centering}p{1.3cm}<{\centering}p{1.3cm}<{\centering}}
		\hline
		\multirow{2}{*}{Backbone} & \multirow{2}{*}{Method} & \multicolumn{3}{c}{1-shot} & \multicolumn{3}{c}{5-shot} \\ \cline{3-8} 
		&             & $\rm mIoU_n$   & $\rm mIoU_b$   & $\rm mIoU_a$  & $\rm mIoU_n$   & $\rm mIoU_b$   & $\rm mIoU_a$  \\ \hline
		\multirow{2}{*}{VGG16}    & BAM ($\rm{w}/\rm{o}$ $\bm{E}$) & 37.54   & \textbf{67.03}   & 59.65  & 41.49   & \textbf{67.03}   & 60.64  \\
		& BAM         & \textbf{43.19}   & \textbf{67.03}   & \textbf{61.07}  & \textbf{46.15}   & 67.02   & \textbf{61.80}  \\ \hline
		\multirow{2}{*}{ResNet50} & BAM ($\rm{w}/\rm{o}$ $\bm{E}$) & 42.37   & \textbf{72.72}   & 65.13  & 43.71   & \textbf{72.72}   & 65.46  \\
		& BAM         & \textbf{47.93}   & \textbf{72.72}   & \textbf{66.52}  & \textbf{49.17}   & \textbf{72.72}   & \textbf{66.83}  \\ \hline
	\end{tabular}}
	\caption{Quantitative results on PASCAL-5$^i$ under generalized FSS setting. ``$\bm E$'' denotes the ensemble module.}
	\label{tab:7}
\end{table*}

\vspace{0.1cm}
\noindent\textbf{Ablation study on two learners.} As mentioned in Sec.\,\ref{sec:4.1}, the two learners could be trained jointly or separately. In our experiments, the latter scheme exhibits better performance, as shown in the first two rows of Tab.\,\ref{tab:4}. We attribute this phenomenon to different utilization of the backbone by the two learners. Specifically, one tends to fix the parameters to enhance generalization, while the other tends to update the parameters to extract more discriminative features, which is challenging to balance in the end-to-end training paradigm. Moreover, we notice that the performance of the model without ${\mathcal{L}_{{\rm{meta}}}}$ becomes slightly worse, indicating the necessity of constraining the prediction results of meta learner.
\vspace{1.0pt}\\
\textbf{Ablation study on ensemble module.} The initial weights of the model have a significant influence on the training process and even the final result. Thus, we conduct relevant ablation studies on this aspect of the ensemble module, which can be regarded as a crucial component of BAM. In our experiments, the ensemble module with initial weights 1 and 0 for meta learner and base learner respectively is markedly superior to other schemes, achieving 2.73\% mIoU improvements over the module with randomly initialized weights, as presented in the 3$^{\rm rd}$\;and 4$^{\rm th}$\;rows of Tab.\,\ref{tab:4}. Furthermore, we also investigate the effect of the adjustment factor $\psi$ on performance, the results of which indicates that adjusting the coarse predictions of the meta learner according to $\psi$ plays an essential role in model ensemble forecasting. Figure\;\ref{fig:5} presents the comparison results between the methods using different low-level features to estimate $\psi$, where the case with ${B_2}$ features shows a better trade-off between segmentation accuracy and computational complexity.
\vspace{1.0pt}\\
\textbf{Ablation study on \textit{K}-shot fusion schemes.} As described in Sec.\,\ref{sec:4.4}, we propose to adaptively adjust the fusion weight of each support sample according to the value of $\psi$. Compared with other solutions, the proposed scheme achieves a sizeable gain (see Tab.\,\ref{tab:5}) under 5-shot setting, further demonstrating the significance of such a factor that measures the differences between images for FSS task. 
\vspace{1.0pt}\\
\textbf{Ablation study on support annotations.} To evaluate the performance of BAM in complex scenarios, we perform experiments with different support annotations. Specifically, in addition to the standard dense mask annotation, bounding box annotation is also introduced for comparison. As can be noticed in Tab.\,\ref{tab:6}, the model with bounding box annotations produces competitive results compared to the model with costly pixel-wise annotations, indicating strong robustness of the proposed scheme. 
\subsection{Generalized Few-Shot Segmentation }
\label{sec:5.4}
In light of the unique nature of the proposed approach, we extend it to a more realistic but challenging setting, \textit{i.e.}, generalized FSS. We simply merge the final output with the output of the base learner according to a predefined threshold $\tau$ to generate the overall segmentation results without any learnable parameters (Eq.\,(\ref{eq:18})). Inspired by the work related to few-shot classification and detection \cite{li2019few,fan2021generalized}, we also define three metrics to evaluate the performance under generalized setting: ${\rm{mIo}}{{\rm{U}}_{\rm{n}}}$, ${\rm{mIo}}{{\rm{U}}_{\rm{b}}}$, and ${\rm{mIo}}{{\rm{U}}_{\rm{a}}}$, denoting the mIoU scores of the novel classes, base classes, and all classes, respectively. As shown in Tab.\,\ref{tab:7}, with the ensemble module, the performance of the segmentation model is enhanced across the board, not just for novel classes. Moreover, the qualitative results in Fig.\,\ref{fig:6} also illustrate its satisfactory capability of handling generalized FSS tasks. 


\section{Conclusion}
We proposed a novel scheme to alleviate the bias problem of FSS models towards the seen concepts. The core idea of our scheme is to leverage the base learner to identify the confusable (base) regions in the query images and further refine the prediction of the meta learner. Surprisingly, even with two plain learners, our scheme also sets new state-of-the-arts on FSS benchmarks. Moreover, we extended the current task to a more challenging generalized setting and produced strong baseline results. We hope that our work could shed light on future research to address the bias or semantic confusion problems. \\

\vspace{-0.55cm}
\section*{Acknowledgements}
\vspace{-0.15cm}
{This work was supported in part by the National Natural Science Foundation of China under Grants 62136007 and U20B2068, and in part by the Shaanxi Science Foundation for Distinguished Young Scholars under Grant 2021JC-16.}

{\small\bibliographystyle{ieee_fullname}
\bibliography{egbib}}

\begin{thebibliography}{10}\itemsep=-1pt

\bibitem{badrinarayanan2017segnet}
Vijay Badrinarayanan, Alex Kendall, and Roberto Cipolla.
\newblock Segnet: A deep convolutional encoder-decoder architecture for image
  segmentation.
\newblock {\em IEEE transactions on pattern analysis and machine intelligence},
  39(12):2481--2495, 2017.

\bibitem{bolya2019yolact++}
Daniel Bolya, Chong Zhou, Fanyi Xiao, and Yong~Jae Lee.
\newblock Yolact++: Better real-time instance segmentation.
\newblock {\em arXiv preprint arXiv:1912.06218}, 2019.

\bibitem{bolya2019yolact}
Daniel Bolya, Chong Zhou, Fanyi Xiao, and Yong~Jae Lee.
\newblock Yolact: Real-time instance segmentation.
\newblock In {\em Proceedings of the IEEE/CVF International Conference on
  Computer Vision}, pages 9157--9166, 2019.

\bibitem{chen2017deeplab}
Liang-Chieh Chen, George Papandreou, Iasonas Kokkinos, Kevin Murphy, and Alan~L
  Yuille.
\newblock Deeplab: Semantic image segmentation with deep convolutional nets,
  atrous convolution, and fully connected crfs.
\newblock {\em IEEE transactions on pattern analysis and machine intelligence},
  40(4):834--848, 2017.

\bibitem{chen2019imageb}
Zitian Chen, Yanwei Fu, Kaiyu Chen, and Yu-Gang Jiang.
\newblock Image block augmentation for one-shot learning.
\newblock In {\em Proceedings of the AAAI Conference on Artificial
  Intelligence}, volume~33, pages 3379--3386, 2019.

\bibitem{chen2019imaged}
Zitian Chen, Yanwei Fu, Yu-Xiong Wang, Lin Ma, Wei Liu, and Martial Hebert.
\newblock Image deformation meta-networks for one-shot learning.
\newblock In {\em Proceedings of the IEEE/CVF Conference on Computer Vision and
  Pattern Recognition}, pages 8680--8689, 2019.

\bibitem{cheng2021task}
Gong Cheng, Ruimin Li, Chunbo Lang, and Junwei Han.
\newblock Task-wise attention guided part complementary learning for few-shot
  image classification.
\newblock {\em Science China Information Sciences}, 64(2):1--14, 2021.

\bibitem{deng2009imagenet}
Jia Deng, Wei Dong, Richard Socher, Li-Jia Li, Kai Li, and Li Fei-Fei.
\newblock Imagenet: A large-scale hierarchical image database.
\newblock In {\em 2009 IEEE conference on computer vision and pattern
  recognition}, pages 248--255. Ieee, 2009.

\bibitem{everingham2010pascal}
Mark Everingham, Luc Van~Gool, Christopher~KI Williams, John Winn, and Andrew
  Zisserman.
\newblock The pascal visual object classes (voc) challenge.
\newblock {\em International journal of computer vision}, 88(2):303--338, 2010.

\bibitem{fan2021generalized}
Zhibo Fan, Yuchen Ma, Zeming Li, and Jian Sun.
\newblock Generalized few-shot object detection without forgetting.
\newblock In {\em Proceedings of the IEEE/CVF Conference on Computer Vision and
  Pattern Recognition}, pages 4527--4536, 2021.

\bibitem{finn2017model}
Chelsea Finn, Pieter Abbeel, and Sergey Levine.
\newblock Model-agnostic meta-learning for fast adaptation of deep networks.
\newblock In {\em International Conference on Machine Learning}, pages
  1126--1135. PMLR, 2017.

\bibitem{gatys2015neural}
Leon~A Gatys, Alexander~S Ecker, and Matthias Bethge.
\newblock A neural algorithm of artistic style.
\newblock {\em arXiv preprint arXiv:1508.06576}, 2015.

\bibitem{gatys2016image}
Leon~A Gatys, Alexander~S Ecker, and Matthias Bethge.
\newblock Image style transfer using convolutional neural networks.
\newblock In {\em Proceedings of the IEEE conference on computer vision and
  pattern recognition}, pages 2414--2423, 2016.

\bibitem{hariharan2011semantic}
Bharath Hariharan, Pablo Arbel{\'a}ez, Lubomir Bourdev, Subhransu Maji, and
  Jitendra Malik.
\newblock Semantic contours from inverse detectors.
\newblock In {\em 2011 International Conference on Computer Vision}, pages
  991--998. IEEE, 2011.

\bibitem{he2017mask}
Kaiming He, Georgia Gkioxari, Piotr Doll{\'a}r, and Ross Girshick.
\newblock Mask r-cnn.
\newblock In {\em Proceedings of the IEEE international conference on computer
  vision}, pages 2961--2969, 2017.

\bibitem{he2016deep}
Kaiming He, Xiangyu Zhang, Shaoqing Ren, and Jian Sun.
\newblock Deep residual learning for image recognition.
\newblock In {\em Proceedings of the IEEE conference on computer vision and
  pattern recognition}, pages 770--778, 2016.

\bibitem{huang2017densely}
Gao Huang, Zhuang Liu, Laurens Van Der~Maaten, and Kilian~Q Weinberger.
\newblock Densely connected convolutional networks.
\newblock In {\em Proceedings of the IEEE conference on computer vision and
  pattern recognition}, pages 4700--4708, 2017.

\bibitem{huang2019ccnet}
Zilong Huang, Xinggang Wang, Lichao Huang, Chang Huang, Yunchao Wei, and Wenyu
  Liu.
\newblock Ccnet: Criss-cross attention for semantic segmentation.
\newblock In {\em Proceedings of the IEEE/CVF International Conference on
  Computer Vision}, pages 603--612, 2019.

\bibitem{jamal2019task}
Muhammad~Abdullah Jamal and Guo-Jun Qi.
\newblock Task agnostic meta-learning for few-shot learning.
\newblock In {\em Proceedings of the IEEE/CVF Conference on Computer Vision and
  Pattern Recognition}, pages 11719--11727, 2019.

\bibitem{jing2019neural}
Yongcheng Jing, Yezhou Yang, Zunlei Feng, Jingwen Ye, Yizhou Yu, and Mingli
  Song.
\newblock Neural style transfer: A review.
\newblock {\em IEEE transactions on visualization and computer graphics},
  26(11):3365--3385, 2019.

\bibitem{lee2020centermask}
Youngwan Lee and Jongyoul Park.
\newblock Centermask: Real-time anchor-free instance segmentation.
\newblock In {\em Proceedings of the IEEE/CVF conference on computer vision and
  pattern recognition}, pages 13906--13915, 2020.

\bibitem{li2019few}
Aoxue Li, Tiange Luo, Tao Xiang, Weiran Huang, and Liwei Wang.
\newblock Few-shot learning with global class representations.
\newblock In {\em Proceedings of the IEEE/CVF International Conference on
  Computer Vision}, pages 9715--9724, 2019.

\bibitem{li2021adaptive}
Gen Li, Varun Jampani, Laura Sevilla-Lara, Deqing Sun, Jonghyun Kim, and
  Joongkyu Kim.
\newblock Adaptive prototype learning and allocation for few-shot segmentation.
\newblock In {\em Proceedings of the IEEE/CVF Conference on Computer Vision and
  Pattern Recognition}, pages 8334--8343, 2021.

\bibitem{li2019finding}
Hongyang Li, David Eigen, Samuel Dodge, Matthew Zeiler, and Xiaogang Wang.
\newblock Finding task-relevant features for few-shot learning by category
  traversal.
\newblock In {\em Proceedings of the IEEE/CVF Conference on Computer Vision and
  Pattern Recognition}, pages 1--10, 2019.

\bibitem{lin2016scribblesup}
Di Lin, Jifeng Dai, Jiaya Jia, Kaiming He, and Jian Sun.
\newblock Scribblesup: Scribble-supervised convolutional networks for semantic
  segmentation.
\newblock In {\em Proceedings of the IEEE conference on computer vision and
  pattern recognition}, pages 3159--3167, 2016.

\bibitem{lin2017refinenet}
Guosheng Lin, Anton Milan, Chunhua Shen, and Ian Reid.
\newblock Refinenet: Multi-path refinement networks for high-resolution
  semantic segmentation.
\newblock In {\em Proceedings of the IEEE conference on computer vision and
  pattern recognition}, pages 1925--1934, 2017.

\bibitem{lin2017feature}
Tsung-Yi Lin, Piotr Doll{\'a}r, Ross Girshick, Kaiming He, Bharath Hariharan,
  and Serge Belongie.
\newblock Feature pyramid networks for object detection.
\newblock In {\em Proceedings of the IEEE conference on computer vision and
  pattern recognition}, pages 2117--2125, 2017.

\bibitem{lin2017focal}
Tsung-Yi Lin, Priya Goyal, Ross Girshick, Kaiming He, and Piotr Doll{\'a}r.
\newblock Focal loss for dense object detection.
\newblock In {\em Proceedings of the IEEE international conference on computer
  vision}, pages 2980--2988, 2017.

\bibitem{lin2014microsoft}
Tsung-Yi Lin, Michael Maire, Serge Belongie, James Hays, Pietro Perona, Deva
  Ramanan, Piotr Doll{\'a}r, and C~Lawrence Zitnick.
\newblock Microsoft coco: Common objects in context.
\newblock In {\em European conference on computer vision}, pages 740--755.
  Springer, 2014.

\bibitem{liu2021anti}
Binghao Liu, Yao Ding, Jianbin Jiao, Xiangyang Ji, and Qixiang Ye.
\newblock Anti-aliasing semantic reconstruction for few-shot semantic
  segmentation.
\newblock In {\em Proceedings of the IEEE/CVF Conference on Computer Vision and
  Pattern Recognition}, pages 9747--9756, 2021.

\bibitem{liu2021harmonic}
Binghao Liu, Jianbin Jiao, and Qixiang Ye.
\newblock Harmonic feature activation for few-shot semantic segmentation.
\newblock {\em IEEE Transactions on Image Processing}, 30:3142--3153, 2021.

\bibitem{liu2020prototype}
Jinlu Liu and Yongqiang Qin.
\newblock Prototype refinement network for few-shot segmentation.
\newblock {\em arXiv preprint arXiv:2002.03579}, 2020.

\bibitem{liu2020crnet}
Weide Liu, Chi Zhang, Guosheng Lin, and Fayao Liu.
\newblock Crnet: Cross-reference networks for few-shot segmentation.
\newblock In {\em Proceedings of the IEEE/CVF Conference on Computer Vision and
  Pattern Recognition}, pages 4165--4173, 2020.

\bibitem{liu2020part}
Yongfei Liu, Xiangyi Zhang, Songyang Zhang, and Xuming He.
\newblock Part-aware prototype network for few-shot semantic segmentation.
\newblock In {\em European Conference on Computer Vision}, pages 142--158.
  Springer, 2020.

\bibitem{long2015fully}
Jonathan Long, Evan Shelhamer, and Trevor Darrell.
\newblock Fully convolutional networks for semantic segmentation.
\newblock In {\em Proceedings of the IEEE conference on computer vision and
  pattern recognition}, pages 3431--3440, 2015.

\bibitem{lu2021simpler}
Zhihe Lu, Sen He, Xiatian Zhu, Li Zhang, Yi-Zhe Song, and Tao Xiang.
\newblock Simpler is better: Few-shot semantic segmentation with classifier
  weight transformer.
\newblock In {\em Proceedings of the IEEE/CVF International Conference on
  Computer Vision}, pages 8741--8750, 2021.

\bibitem{min2021hypercorrelation}
Juhong Min, Dahyun Kang, and Minsu Cho.
\newblock Hypercorrelation squeeze for few-shot segmentation.
\newblock {\em arXiv preprint arXiv:2104.01538}, 2021.

\bibitem{nguyen2019feature}
Khoi Nguyen and Sinisa Todorovic.
\newblock Feature weighting and boosting for few-shot segmentation.
\newblock In {\em Proceedings of the IEEE/CVF International Conference on
  Computer Vision}, pages 622--631, 2019.

\bibitem{nichol2018first}
Alex Nichol, Joshua Achiam, and John Schulman.
\newblock On first-order meta-learning algorithms.
\newblock {\em arXiv preprint arXiv:1803.02999}, 2018.

\bibitem{peng2017large}
Chao Peng, Xiangyu Zhang, Gang Yu, Guiming Luo, and Jian Sun.
\newblock Large kernel matters--improve semantic segmentation by global
  convolutional network.
\newblock In {\em Proceedings of the IEEE conference on computer vision and
  pattern recognition}, pages 4353--4361, 2017.

\bibitem{rakelly2018conditional}
Kate Rakelly, Evan Shelhamer, Trevor Darrell, Alyosha Efros, and Sergey Levine.
\newblock Conditional networks for few-shot semantic segmentation.
\newblock 2018.

\bibitem{ravi2016optimization}
Sachin Ravi and Hugo Larochelle.
\newblock Optimization as a model for few-shot learning.
\newblock 2016.

\bibitem{redmon2016you}
Joseph Redmon, Santosh Divvala, Ross Girshick, and Ali Farhadi.
\newblock You only look once: Unified, real-time object detection.
\newblock In {\em Proceedings of the IEEE conference on computer vision and
  pattern recognition}, pages 779--788, 2016.

\bibitem{ren2015faster}
Shaoqing Ren, Kaiming He, Ross Girshick, and Jian Sun.
\newblock Faster r-cnn: Towards real-time object detection with region proposal
  networks.
\newblock {\em Advances in neural information processing systems}, 28:91--99,
  2015.

\bibitem{ronneberger2015u}
Olaf Ronneberger, Philipp Fischer, and Thomas Brox.
\newblock U-net: Convolutional networks for biomedical image segmentation.
\newblock In {\em International Conference on Medical image computing and
  computer-assisted intervention}, pages 234--241. Springer, 2015.

\bibitem{shaban2017one}
Amirreza Shaban, Shray Bansal, Zhen Liu, Irfan Essa, and Byron Boots.
\newblock One-shot learning for semantic segmentation.
\newblock {\em arXiv preprint arXiv:1709.03410}, 2017.

\bibitem{siam2019amp}
Mennatullah Siam, Boris~N Oreshkin, and Martin Jagersand.
\newblock Amp: Adaptive masked proxies for few-shot segmentation.
\newblock In {\em Proceedings of the IEEE/CVF International Conference on
  Computer Vision}, pages 5249--5258, 2019.

\bibitem{simonyan2014very}
Karen Simonyan and Andrew Zisserman.
\newblock Very deep convolutional networks for large-scale image recognition.
\newblock {\em arXiv preprint arXiv:1409.1556}, 2014.

\bibitem{snell2017prototypical}
Jake Snell, Kevin Swersky, and Richard~S Zemel.
\newblock Prototypical networks for few-shot learning.
\newblock {\em arXiv preprint arXiv:1703.05175}, 2017.

\bibitem{sung2018learning}
Flood Sung, Yongxin Yang, Li Zhang, Tao Xiang, Philip~HS Torr, and Timothy~M
  Hospedales.
\newblock Learning to compare: Relation network for few-shot learning.
\newblock In {\em Proceedings of the IEEE conference on computer vision and
  pattern recognition}, pages 1199--1208, 2018.

\bibitem{tian2020prior}
Zhuotao Tian, Hengshuang Zhao, Michelle Shu, Zhicheng Yang, Ruiyu Li, and Jiaya
  Jia.
\newblock Prior guided feature enrichment network for few-shot segmentation.
\newblock {\em IEEE Transactions on Pattern Analysis \& Machine Intelligence},
  (01):1--1, 2020.

\bibitem{vanschoren2018meta}
Joaquin Vanschoren.
\newblock Meta-learning: A survey.
\newblock {\em arXiv preprint arXiv:1810.03548}, 2018.

\bibitem{vilalta2002perspective}
Ricardo Vilalta and Youssef Drissi.
\newblock A perspective view and survey of meta-learning.
\newblock {\em Artificial intelligence review}, 18(2):77--95, 2002.

\bibitem{vinyals2016matching}
Oriol Vinyals, Charles Blundell, Timothy Lillicrap, Daan Wierstra, et~al.
\newblock Matching networks for one shot learning.
\newblock {\em Advances in neural information processing systems},
  29:3630--3638, 2016.

\bibitem{wang2020few}
Haochen Wang, Xudong Zhang, Yutao Hu, Yandan Yang, Xianbin Cao, and Xiantong
  Zhen.
\newblock Few-shot semantic segmentation with democratic attention networks.
\newblock In {\em Computer Vision--ECCV 2020: 16th European Conference,
  Glasgow, UK, August 23--28, 2020, Proceedings, Part XIII 16}, pages 730--746.
  Springer, 2020.

\bibitem{wang2019panet}
Kaixin Wang, Jun~Hao Liew, Yingtian Zou, Daquan Zhou, and Jiashi Feng.
\newblock Panet: Few-shot image semantic segmentation with prototype alignment.
\newblock In {\em Proceedings of the IEEE/CVF International Conference on
  Computer Vision}, pages 9197--9206, 2019.

\bibitem{wang2016learning}
Yu-Xiong Wang and Martial Hebert.
\newblock Learning to learn: Model regression networks for easy small sample
  learning.
\newblock In {\em European Conference on Computer Vision}, pages 616--634.
  Springer, 2016.

\bibitem{wu2021learning}
Zhonghua Wu, Xiangxi Shi, Guosheng Lin, and Jianfei Cai.
\newblock Learning meta-class memory for few-shot semantic segmentation.
\newblock In {\em Proceedings of the IEEE/CVF International Conference on
  Computer Vision}, pages 517--526, 2021.

\bibitem{xie2020polarmask}
Enze Xie, Peize Sun, Xiaoge Song, Wenhai Wang, Xuebo Liu, Ding Liang, Chunhua
  Shen, and Ping Luo.
\newblock Polarmask: Single shot instance segmentation with polar
  representation.
\newblock In {\em Proceedings of the IEEE/CVF conference on computer vision and
  pattern recognition}, pages 12193--12202, 2020.

\bibitem{xie2021scale}
Guo-Sen Xie, Jie Liu, Huan Xiong, and Ling Shao.
\newblock Scale-aware graph neural network for few-shot semantic segmentation.
\newblock In {\em Proceedings of the IEEE/CVF Conference on Computer Vision and
  Pattern Recognition}, pages 5475--5484, 2021.

\bibitem{yang2020prototype}
Boyu Yang, Chang Liu, Bohao Li, Jianbin Jiao, and Qixiang Ye.
\newblock Prototype mixture models for few-shot semantic segmentation.
\newblock In {\em European Conference on Computer Vision}, pages 763--778.
  Springer, 2020.

\bibitem{yang2021mining}
Lihe Yang, Wei Zhuo, Lei Qi, Yinghuan Shi, and Yang Gao.
\newblock Mining latent classes for few-shot segmentation.
\newblock {\em arXiv preprint arXiv:2103.15402}, 2021.

\bibitem{yu2015multi}
Fisher Yu and Vladlen Koltun.
\newblock Multi-scale context aggregation by dilated convolutions.
\newblock {\em arXiv preprint arXiv:1511.07122}, 2015.

\bibitem{zhang2021self}
Bingfeng Zhang, Jimin Xiao, and Terry Qin.
\newblock Self-guided and cross-guided learning for few-shot segmentation.
\newblock In {\em Proceedings of the IEEE/CVF Conference on Computer Vision and
  Pattern Recognition}, pages 8312--8321, 2021.

\bibitem{zhang2019pyramid}
Chi Zhang, Guosheng Lin, Fayao Liu, Jiushuang Guo, Qingyao Wu, and Rui Yao.
\newblock Pyramid graph networks with connection attentions for region-based
  one-shot semantic segmentation.
\newblock In {\em Proceedings of the IEEE/CVF International Conference on
  Computer Vision}, pages 9587--9595, 2019.

\bibitem{zhang2019canet}
Chi Zhang, Guosheng Lin, Fayao Liu, Rui Yao, and Chunhua Shen.
\newblock Canet: Class-agnostic segmentation networks with iterative refinement
  and attentive few-shot learning.
\newblock In {\em Proceedings of the IEEE/CVF Conference on Computer Vision and
  Pattern Recognition}, pages 5217--5226, 2019.

\bibitem{zhang2018sg}
Xiaolin Zhang, Yunchao Wei, Yi Yang, and Thomas~S Huang.
\newblock Sg-one: Similarity guidance network for one-shot semantic
  segmentation.
\newblock {\em arXiv preprint arXiv:1810.09091}, 2018.

\bibitem{zhao2017pyramid}
Hengshuang Zhao, Jianping Shi, Xiaojuan Qi, Xiaogang Wang, and Jiaya Jia.
\newblock Pyramid scene parsing network.
\newblock In {\em Proceedings of the IEEE conference on computer vision and
  pattern recognition}, pages 2881--2890, 2017.

\end{thebibliography}

\clearpage
\setcounter{equation}{0}
\setcounter{figure}{0}
\setcounter{table}{0}
\setcounter{page}{1}
\renewcommand\thefigure{S\arabic{figure}}
\renewcommand\thetable{S\arabic{table}}
\renewcommand{\theequation}{S\arabic{equation}}
\twocolumn[
\begin{@twocolumnfalse}
	\section*{\centering{\Large Supplementary Material for \\ \emph{Learning What Not to Segment: A New Perspective on Few-Shot Segmentation\\[15pt]}}}
\end{@twocolumnfalse}
]
\begin{spacing}{1.5}{\noindent\textbf{\fontsize{13.0pt}{\baselineskip}\selectfont A. Calculation of FLOPs}}\end{spacing}
\indent Floating point operations per second (FLOPs) is utilized to evaluate the computational complexity of our model in the ablation study (refer to Sec.\,\ref{sec:5.3}). Here we introduce the specific calculation process. Given the low-level features ${\bf{f}}_{{\rm{low}}}^{\rm{s}},{\bf{f}}_{{\rm{low}}}^{\rm{q}} \in \mathbb{R}{^{C \times H \times W}}$, we first compute the corresponding Gram matrices ${{\bf{G}}^{\rm{s}}},{{\bf{G}}^{\rm{q}}}$. Second, we perform subtraction ${{\bf{G}}^{\rm{s}}} - {{\bf{G}}^{\rm{q}}}$ to evaluate the difference. Finally, we calculate the Frobenius norm of the difference to get an overall indicator $\psi$. The number of FLOPs can be defined as:
\begin{equation}
\begin{array}{l}
{\rm{FLOPs}} = 4{C^2}N + {C^2} + 2{C^2}\\
{\kern 3.1em}= {C^2}\left( {4N + 3} \right)
\end{array},
\end{equation}
where $N = H \times W$, and the three terms in the first row correspond to the three processes above\footnote{Taking the third convolutional block ${B_2}$ of ResNet50 \cite{he2016deep} backbone as an example, the FLOPs is 3.78G with ${{\bf{f}}_{{\rm{low}}}} \in \mathbb{R}{^{512 \times 60 \times 60}}$.}.
\vspace{0.1cm}
\begin{spacing}{1.5}{\noindent\textbf{\fontsize{13.0pt}{\baselineskip}\selectfont B. Implementation Details}}\end{spacing}
\indent As mentioned in Sec.\,\ref{sec:4.5}, we simply fuse the predictions of the base learner and the final predictions after ensemble according to a predefined threshold $\tau$$=$$0.9$ to obtain the holistic segmentation results ${{\bf{\hat m}}_{\rm{g}}}$, as presented in Eq.\,(\ref{eq:18}). There is actually another alternative extension scheme, which can be defined as:
\begin{equation}
{\bf{\hat m}}_{\rm{g}}^{(x,y)}{\rm{ = }}\left\{ \begin{array}{l}
1{\kern 2.8em}{\bf{p}}_{\rm{f}}^{1;\left( {x,y} \right)} > \tau \;{\rm{and }}\;{\bf{\hat m}}_{\rm{b}}^{(x,y)}{\rm{ = 0}}\\
{\bf{\hat m}}_{\rm{b}}^{(x,y)}{\kern 1.2ex}{\bf{\hat m}}_{\rm{b}}^{(x,y)} \ne {\rm{0}}\\
0{\kern 2.8em}{\rm{otherwise}}
\end{array} \right..
\end{equation}
The differences between two schemes are as follows: the former is mainly based on the final predictions, leveraging the base learner to determine the base pixels in the background region; the latter, by contrast, is primarily based on the predictions of the base learner, using the final predictions to determine the novel pixels in the background region. In our experiments, BAM achieves similar results using both schemes, with the former slightly better; however, the baseline approach (w/o ensemble module) can only produce tolerable results when the latter scheme is adopted, which further verifies the importance of the ensemble module to correct the coarse predictions of meta learners.\vspace{5pt}\\

\end{document}